%% file: Basilico-De_Nittis-Gatti_-_Multi-resource_defensive_strategies_for_patrolling_games_with_alarm_systems.tex
\documentclass{article}
\usepackage{ijcai16}
\usepackage{times}
\usepackage{helvet}
\usepackage{courier}
\usepackage{amssymb}
\usepackage{amsmath}
\usepackage{array}
\usepackage{float}
\usepackage{graphics,graphicx,epsfig,epstopdf}
\usepackage{psfrag}
\usepackage{subfigure}
\usepackage{tikz}
\usepackage{url}
\usepackage{color}
\usepackage{theorem}
\usepackage{algorithm}
\usepackage[noend]{algpseudocode}

\usepackage{verbatim}
\usepackage{enumitem}
\usepackage{multirow}
\usepackage{mathabx}
\usepackage{xr}

\newtheorem{theorem}{Theorem}

\newtheorem{lemma}[theorem]{Lemma}
\newtheorem{definition}{Definition}

\newcommand{\Att}{$\mathcal{A}$}
\newcommand{\Def}{$\mathcal{D}$}

\title{Multi--resource defensive strategies for patrolling games with alarm systems}

\author{Nicola Basilico \\
Dept. of Computer Science \\
University of Milan \\
Milano, Italy \\
nicola.basilico@unimi.it
\And
Giuseppe De Nittis and Nicola Gatti \\
Dept. of Electronics, Information and Bioengineering \\
Politecnico di Milano \\
Milano, Italy \\
\{giuseppe.denittis, nicola.gatti\}@polimi.it}

\begin{document}
\maketitle

\begin{abstract}
Security Games employ game theoretical tools to derive resource allocation strategies in security domains. Recent works considered the presence of alarm systems, even suffering various forms of uncertainty, and showed that disregarding alarm signals may lead to arbitrarily bad strategies. The central problem with an alarm system, unexplored in other Security Games, is finding the best strategy to respond to alarm signals for each mobile defensive resource. The literature provides results for the basic single--resource case, showing that even in that case the problem is computationally hard. In this paper, we focus on the challenging problem of designing algorithms scaling with multiple resources. 
First, we focus on finding the minimum number of resources assuring non--null protection to every target. Then, we deal with the computation of multi--resource strategies with different degrees of coordination among resources. For each considered problem, we provide a computational analysis and propose algorithmic methods.
\end{abstract}

\input{01-introduction}
\input{02-problem}
\input{03-minimum_resources}
\input{04-signal_response_game}
\input{05-covering_placement}
\input{06-experiments}
\input{07-conclusions}

\bibliographystyle{named}
\bibliography{references}

\end{document}

%% file: 01-introduction.tex
\section{Introduction}
Security Games represent one of the most successful application of non--cooperative game theory in the real world~\cite{jain2012overview}. The basic approach is to model a security situation with a 2--player game, between a \emph{Defender} and an \emph{Attacker}, and to derive the best strategy of the Defender, most of the times according to the Stackelberg paradigm. A wide literature on Security Games studies issues like resource scheduling constraints~\cite{DBLP:conf/atal/KiekintveldJTPOT09}, bounded rationality~\cite{yang2011improving}, Attacker's observation models~\cite{an2013security}, protection of infrastructures~\cite{BlumAAAI14}. 

Real security systems are equipped with \emph{sensors} that are capable to trigger \emph{alarms} when attacks are detected but may suffer from various forms of uncertainty. Mobile defensive resources (a.k.a. \emph{patrollers}) can respond to alarm signals covering the targets potentially under attack. It is demonstrated that disregarding alarm signals may lead to strategies arbitrarily worse than those obtained when alarm signals are exploited. Nevertheless, the study of how to include alarm systems in Security Games is largely unexplored and represents one of the most challenging open problems in the field. In particular, the central question is: \emph{given an alarm signal, how should the Defender respond to it at best?} \cite{bdg2015advpatr} study the scenario with only one resource available to the Defender and with sensors suffering from \emph{spatial uncertainty}, i.e., an alarm signal is raised whenever an attack is performed, but the Defender is uncertain on the actual attacked location, as in border patrolling~\cite{gal2008border}. In such situations, the best strategy is to stay in a location, wait for an alarm signal and then respond to it at best. This last task is proved $\mathcal{APX}$--hard. \cite{debenignis2016} study the scenario with sensors suffering both from \emph{spatial uncertainty} and \emph{false negatives} when only one resource is available to the Defender, showing that it is $\mathcal{PSPACE}$--hard. For small missed detection rates, placing in a location and responding to an alarm signal keeps to be the best strategy for the Defender, while for large missed detection rates the best strategy is to patrol a number of targets even in absence of alarm signals and respond to an alarm signal once triggered.

{\bf Original contributions}. In this paper, we focus on settings with a spatial uncertain alarm system and multiple defensive resources. Here the challenge is to design algorithms able to scale with the resources.
We provide some original contributions. In real--world settings, due to budget constraints, it is important to minimize the number of resources assuring a minimum protection level. We study the problem of finding the minimum number of resources assuring non--null protection to every target. 

This number corresponds to the lower bound over the number of resources one should employ in practice. We show that the problem is log--$\mathcal{APX}$--complete on general graphs, while it is in $\mathcal{P}$ for special graphs, such as trees and cycles, that are rather usual in real--world applications.
Then, we study the problem of finding the best strategy to respond to any alarm signal once a disposition of resources in the environment is given, according to different \emph{degrees of coordination} among the resources. Finally, we propose an approach that includes the previous algorithms and that returns in anytime fashion the best disposition of resources. For each of our algorithms, we provide a thorough experimental evaluation showing that they can solve realistic settings.

%% file: 02-problem.tex
\section{Security game model and previous results}
Our security game is a generalization of~\cite{basilico2015security}, obtained allowing the Defender to control an arbitrary number of $m$ resources instead of just a single one. We summarize its main features. A {\em patrolling setting} is a graph $G=(V,E)$ representing an environment where areas that can be attacked are given by $n$ target vertices, denoted as $T \subseteq V$. A target $t\in T$ has a value $\pi(t)\in (0,1]$ and requires $d(t)\in \mathbb{N}^+$ time units ({\em penetration time}) to be compromised. Edges in $E$ are unitary, while $\omega^*_{i,j}$ is the smallest traveling cost beweem vertices $i$ and $j$. An \emph{alarm system} $(S, p)$ generates a signal $s \in S$ if target $t$ is attacked with probability $p(s \mid t)$. $S$, $p$ and any generated signal $s$ are public knowledge. We call $T(s)=\{t \in T \mid p(s \mid t) > 0\}$ the {\em support of a signal} $s$ and $S(t)=\{s \in S \mid p(s \mid t) > 0\}$ the {\em support of a target} $t$. We assume that $p(\emptyset \mid t) = p(s \mid \emptyset) = 0$ for any $t \in T$ and $s \in S$, i.e., there are no false positives or missed detections, and that $|S(t)| \ge 1$ for any $t \in T$.

A 2--player security game takes place between an Attacker \Att~and a Defender \Def. In this game, \Att~seeks to gain value by compromising some target while \Def~can control $m$ defending resources by specifying a movement strategy for them. At any turn of the game, \Att~and \Def~play simultaneously: $\mathcal{A}$ observes the position of the $m$ and decides whether to attack a target (we assume that \Att~can instantly reach the attacked target, this can be relaxed as shown in~\cite{BasilicoGatti2009capturing}.) or to wait. On the opposite side, \Def~has no information about \Att~but, if an attack is present, it observes the alarm signal and decides where to allocate the $m$ resources. Each resource can be allocated either to the same vertex or to an adjacent one. If \Att~attacks target $t$ and a resource is allocated on $t$ in any of the next $d(t)$ turns, then \Def~and \Att~receive payoffs of $1$ and $0$, respectively. Otherwise, they receive $1-\pi(t)$ and $\pi(t)$. A route $r$ for a single resource is a sequence of targets to which the resource is moved where $l_r$ is the route length, $r(i)$ is the $i$--th assigned target and $A(r(i)) = \sum_{h=1}^{i-1}\omega^*_{r(h),r(h+1)}$ is the time between visits to $r(1)$ and $r(i)$. We say that $r_{s,j}$ is a {\em covering} route for resource $j$ under signal $s$ if, $\forall i \in \{1, \ldots, l_r\}$, we have that $r(i) \in T(s)$ and $A(r(i)) \leq d(r(i))$. The interpretation is that when $s$ is generated, having a resource following $r$ will guarantee protection over $T(s)$. A {\em joint} covering route is an $m$--tuple $r_s = \langle r_{s,1}, r_{s,2}, \ldots, r_{s,m} \rangle$. To simplify the notation, we will omit signal $s$ when clear from the context.

The resolution approach for $m=1$ is given in~\cite{basilico2015security}. With no false positives and missed detections, the best strategy for the patroller is to stay on a vertex, wait for a signal, and then respond to it at best. The same holds with multiple resources (the extension of the proof is straightforward) and allows us to decompose the problem into two phases: determining the best signal response strategy from every vertex (\emph{Signal Response Game}), and then determining the best vertex in which the patroller can place (\emph{Patrolling Game}). Once a signal $s$ is received, \Def~has to respond by randomizing over the possible covering routes under signal $s$ and starting from $v$. The signal response strategy is defined as $\sigma^{\mathcal{D}}_{v}=(\sigma^{\mathcal{D}}_{v,s_1},\sigma^{\mathcal{D}}_{v,s_2}, \ldots)$, that is a response strategy for every possible signal. Despite the game being constant--sum, the problem is shown to be $\mathcal{APX}$--{\em hard}, its difficulty mainly being in generating the covering routes under a signal $s$ and starting from $v$. An approximation algorithm for such difficult subproblem is given in~\cite{bdg2015advpatr}. With multiple resources, several questions, related to finding the best number of resources to employ, their placement, and their signal response strategies, arise. In the next sections, we provide some answers to them.

%% file: 03-minimum_resources.tex
\section{Minimizing the number of resources}
Ideally, the determination of the best number of defensive resources must take into account both the level of protection that can be achieved and the costs of additional resources. The minimum protection level a Defender should guarantee is that for each target $t \in T$ there is at least a resource in a vertex $v$ with $\omega_{v,t}^*\leq d(t)$, such that it can cover $t$ by $d(t)$, thus stopping a potential ongoing attack. The least number of resources satisfying this requirement represents the lower bound necessary for the protection of any environment, while the upper bound is the number of resources such that, for any signal $s$, there is a response strategy to $s$ covering all the targets in $T(s)$ thus assuring the Defender a utility of 1. Here we study the problem of computing the minimum number of resources guaranteeing each target a non--null protection. We start by defining the concept of covering placement.

\begin{definition}[Covering Placement]\label{def:cov_placement} A covering placement is defined as a set of vertices $P = \{p_1, \ldots, p_m\}$ where $p_i \in V$, $p_i \neq p_j$ if $i\neq j$ and for every target $t\in T$ there exists a $p_i \in P$ such that $\omega^*_{p_i,t} \le d(t)$.
\end{definition}

\subsection{Arbitrary instances}
With arbitrary graphs, we state the following results.

\begin{theorem}\label{thm:min_resources_hardness}
Computing the minimum covering placement is log--$\mathcal{APX}$--hard even in the basic case all the vertices are targets with penetration times equal to 1.
\end{theorem}

\noindent
\emph{Proof}. Log--$\mathcal{APX}$--hardness of our problem is direct from DOMINATING--SET~\cite{escoffier2006completeness} that is known to be log--$\mathcal{APX}$--complete. This latter problem is defined as follows: given a graph $\overline{G}=(\overline{V},\overline{E})$, minimize $|\overline{V}'|$ with $\overline{V}' \subseteq \overline{V}$ with the property that for all $v \in \overline{V} \setminus \overline{V}'$ there is at least a $v' \in \overline{V}'$ such that $(v,v') \in \overline{E}$. Every instance of DOMINATING--SET with graph $\overline{G}$ is equivalent to the problem of minimizing the covering placement over graph $\overline{G}$ in which all the vertices $\overline{V}$ are targets and have $d = 1$.\hfill$\Box$

\begin{theorem}\label{thm:min_resources_completeness}
Computing the minimum covering placement is in log--$\mathcal{APX}$.
\end{theorem}

\noindent
\emph{Proof}. The membership to log--$\mathcal{APX}$ follows from the fact that every instance of our problem can be formulated in direct way as a SET--COVER instance and that SET--COVER in is in log--$\mathcal{APX}$~\cite{chvatal1979greedy}---our problem is more general than DOMINATING--SET, this justifies the need to consider SET--COVER. More precisely, given an instance of our problem, for each vertex $v \in V$ we can find in polynomial time the set of targets $t$ such that $\omega_{v,t}^*\leq d(t)$. Thus, we map set $T$ to the universe set of SET--COVER and each vertex $v$ to a set in SET--COVER. The objective functions of the two problems are the same. Therefore, every $\alpha$ approximation of SET--COVER is also an $\alpha$ approximation of our problem. \hfill$\Box$

The proof of Theorem~\ref{thm:min_resources_completeness} is based on the fact that every instance of our problem can be formulated in direct way as a SET--COVER instance. As a consequence we can find a solution to our problem by formulating it as a SET--COVER and then by using algorithms for this latter problem: integer linear programming (ILP) for finding the exact solution and greedy algorithms to find an approximate solution~\cite{chvatal1979greedy}. In our approach, we also apply a local search~\cite{musliu2006local} to the solution generated by the greedy algorithm to improve its quality.

\subsection{Special instances: tree and cycle graphs}
\label{sec:special_instances}
Interestingly, in particular instances that are rather usual in real--world applications the problem is optimally solvable in polynomial time. Let us start from the following lemma.

\input{algorithm_trees}

\begin{lemma}
A minimum covering placement in a tree rooted in $\hat{v}$ can be computed in polynomial time with Algorithm~\ref{alg:tree} by calling $TM(\hat{v})$.
\end{lemma}

Algorithm~\ref{alg:tree} works recursively taking as input a vertex $v \in V$. To ease its description, let us assume that $T = V$. The case including non--target vertices only amounts to minor modifications. Binary variables $res_v$ are initially set to $0$ and their assertion corresponds to place a resource on $v \in V$. With $Succ(v) \subseteq V$ we denote all the immediate successors of $v$ on the path leaving the root. The idea is to recursively allocate resources by processing the graph in a bottom--up fashion, from its leaves to the root. Let us consider a function call for a generic vertex $v$. By recursively invoking $TM(v')$ for each $v' \in Succ(v)$ we get, for each successor, a {\em coverage profile} defined with a pair of values $\big(Cov(v'), UnCov(v')\big)$. They encode the following conditions under the currently built resource placement. If variables $res_\nu$ of the subtree rooted in $v'$ constitute a coverage placement for whole the subtree, the coverage profile is such that $Cov(v') = k<\infty$ where $k$ is the distance between $v$ and the closest resource on such subtree and $UnCov(v') = \infty$. Otherwise, the coverage profile is such that $Cov(v') = \infty$ and $UnCov(v') = k < \infty$ where $k$ is the distance from $v$ by which we need to place a resource to have a coverage placement for the subtree rooted in $v'$.

We start from an empty placement and derive coverage profiles recursively from the base case in which $v$ is a leaf (Line~\ref{alg:basecase}). Since deadlines are non--null and costs are unitary, a resource on a leaf is never necessary: we can always cover it from any ancestor whose distance from $v$ is $\le d(v)-1$. Hence the coverage profile for the base case is $(\infty, d(v)-1)$ (Line~\ref{alg:basecase_ret}).

Let us consider the recursive step in which $v$ is a non--leaf vertex. From Line~\ref{alg:recstep} we have all the coverage profiles of each one of the immediate successors of $v$ and we must return the coverage profile of $v$ and decide if a resource must be placed on it. We have three cases.
\begin{itemize}
\item {\em Case~1 (Lines~\ref{alg:recstep}--\ref{alg:case1})}: if the condition of Line~\ref{alg:recstep} is satisfied then the subtree rooted at $v$ is covered by the resources we placed in such subtree. Hence we do not need any resource in $v$. Moreover, the father of $v$ will be at distance $\min\limits_{v'} Cov(v') + 1$ from the closest of such resources.
\item {\em Case~2 (Lines~\ref{alg:case2a}--\ref{alg:case2b}}): if the condition of Line~\ref{alg:case2a} is satisfied then the subtree rooted at $v$ is {\em not} covered by the resources we placed in such subtree. We can achieve coverage by placing a resource in $v$ or in any ancestor whose distance from $v$ should not exceed $\min\limits_{v'} \{ UnCov(v'), d(v)\} - 1$. Since we are trying to minimize the number of resources, there is always interest in postponing a resource allocation in our bottom--up processing of the tree. Therefore, we do not allocate a resource in $v$ and we return the strategy profile naturally resulting from the above considerations (Line~\ref{alg:case2b}).
\item {\em Case~3 (Lines~\ref{alg:case3a}--\ref{alg:case3b}}): if none of the previous two cases is verified, then no resource in any ancestor of $v$ can complete the coverage for the subtree rooted at $v$. We are forced to place a resource in $v$ which makes the associated coverage profile equal to $(1, \infty)$ (Line~\ref{alg:case3b}).
\end{itemize}

Algorithm~\ref{alg:tree} can be adopted also to solve cycle graphs by extracting the $n$ linear subgraphs spanning all the $n$ targets, solving each of them with Algorithm~\ref{alg:tree}, and then selecting the solution with the least number of resources. We summarize the above positive results in the following theorem.

\begin{theorem}
Given $k$, deciding if there exists a covering placement with size $m \le k$ is $\mathcal P$ for trees and cycle graphs.
\end{theorem}

%% file: algorithm_trees.tex
\begin{algorithm}\caption{TM($v$)}\label{alg:tree}
\begin{scriptsize}
\begin{algorithmic}[1]
\If {$v$ is a leaf} \label{alg:basecase}
	\State \Return $(\infty, d(v)-1)$ \label{alg:basecase_ret}
\Else
	\ForAll{$v' \in Succ(v)$}
		\State $\big(Cov(v'), UnCov(v')\big) \gets TM\big(v'\big)$
	\EndFor 

	\If{$\min\limits_{v'} \{ UnCov(v'), d(v)\} - \min\limits_{v'} Cov(v') \geq 0$} \label{alg:recstep}
		\State \Return $(\min\limits_{v'} Cov(v') + 1 , \infty)$ \label{alg:case1}
	\ElsIf{$\min\limits_{v'} \{ UnCov(v'), d(v)\} - 1 \geq 0$} \label{alg:case2a}
		\State \Return $(\infty, \min\limits_{v'} \{ UnCov(v'), d(v)\} - 1)$ \label{alg:case2b}
	\Else 
		\State $res_v \gets 1$ \label{alg:case3a} 
		\State \Return $(1, \infty )$ \label{alg:case3b}
	\EndIf
\EndIf
\end{algorithmic}
\end{scriptsize}
\end{algorithm}

%% file: 04-signal_response_game.tex
\section{Signal response}
In this section, we deal with the problem of how to compute a signal response strategy for the $m$ resources once we are given a joint placement $P \subseteq V^m$, which satisfies the coverage requirement (see Definition~\ref{def:cov_placement}), and each of the $m$ necessary resources is placed at $p_i \in P$. To ease the presentation, we assume that only one signal $s$ is present in $S$ and that $T(s) = T$, omitting $s$ in our formulas. The general case with multiple signals can be obtained by simply refining notation. The extra computational cost due to signals is linear in their number. 

In general, any resource~$i$ will always move along a covering route. Let us denote with $R_i$ the set of covering routes for resource $i$ starting from $p_i$ (we will omit the dependency on $p_i$ since $P$ is always fixed in the scope of a signal response game). In order to compute the covering routes, we resort the exact and approximate methods proposed in~\cite{bdg2015advpatr}.
The presence of multiple resources introduces the coordination dimension~\cite{BasilicoGattiAAAI2010}. We consider three coordination schemes and define three \textit{Signal Response Oracles} (SRO). Each oracle works on a different game model and returns the signal response strategy from a given joint placement $P$ under the corresponding scheme. Strategies for \Def~and \Att~are denoted as $\sigma^{\mathcal{D}}$ and $\sigma^{\mathcal{A}}$, respectively. If not defined differently, $\sigma^{\mathcal{A}} : T \rightarrow [0,1]$ gives the probability $\sigma^{\mathcal{A}}(t)$ of attacking a target $t$. 

\subsection{Full coordination SRO (FC--SRO)}
In the full coordination scheme, we assume that \Def~acts as central planner and executor of the signal response strategy. In other words, \Def~has full control over the resources at any time of the game. It computes a joint strategy for them and jointly moves them on the graph according to it. This scheme can describe scenarios where resources are connected to a central control unit from where orders are issued (e.g., police patrols equipped with radio transceivers). The security game originating from this scheme is played by \Def~and \Att~where $\sigma^{\mathcal{D}}: R \rightarrow [0,1]$, $R=\bigtimes_{i=1}^{m}R_i$ and $\sigma^{\mathcal{D}}(r)$ is the probability of playing the joint covering route $r$. We define $I(r,t)$ as a function returning $1$ if and only if target $t$ is protected by $r$ (we allow $r$ to be both a joint and a single--resource covering route) and $0$ otherwise. For any $r \in R$ and $t \in T$, $(r,t)$ is a game outcome where \Def~and \Att~receive $U_{\mathcal{D}}(r,t)= 1 - (1 - I(r,t))\pi(t)$ and $U_{\mathcal{A}}(r,t) = 1-U_{\mathcal{D}}(r,t)$, respectively. The game is constant--sum and can be solved computing the maxmin strategy with a linear program. However, each $R_i$ can have exponential size and its computation entails the resolution of an $\mathcal{NP}$--{\em hard} problem. Even adopting the approximated method and working with incomplete sets of covering routes would require us to enumerate over all the possible combinations to construct the space of joint covering routes, which are exponentially many in the number of resources. However, we observe that in our case there always exists a minmax equilibrium where \Def~plays on at most $|T|$ routes as shown in from~\cite{supportopiccolo}.

\begin{theorem}
Given a game with $K$ players in which all the players $k \in K \setminus\{1\}$ have $m$ actions and player $1$ has $n$ actions, there is a minmax strategy of players $K \setminus\{1\}$ in which the support of each player has a size not larger than $n$.
\end{theorem}

For this reason, we devise a row--generation approach that iteratively generates rows in the game matrix (joint covering routes). The row generation routine first generates, for each resource $i$, a set of covering routes $R_i$. Then, it considers a joint covering route $r'$ ({\em initial} covering route). Set $R = \{r'\}$ and solve the corresponding constant--sum game using $R$ as the action space for \Def. Solve the following ILP where $x_{ir}$ is a binary variable taking value of $1$ when route $r \in R_i$ is selected for resource $i$, $y_t$ is a binary variable taking value of $1$ when target $t$ is protected by the set of selected covering routes, and $\sigma^{\mathcal{A}}$ is \Att's minmax strategy in the previously solved game.

\begin{scriptsize}
\begin{align*}
\arg\max \hspace{0.5cm} 1 - \sum_{t \in T}\sigma^{\mathcal{A}}(t) \pi(t) (1 - y_t) & 	\quad \text{s.t.} \\
\sum_{i=1}^{m} \sum_{r \in R_i} I(r,t) x_{ir}-y_t \geq 0 & \quad  \forall t \in T \\
\sum_{r \in R_i} x_{ir}=1 & \quad \forall i \in \{1 \ldots m\}
\end{align*}
\end{scriptsize}

From the solution of this problem we obtain a joint covering route $\hat{r}$ where resource $i$ plays route $i$ if $x_{ri}=1$. Clearly, $\hat{r}$ is a best response for \Def~in the incumbent game equilibrium. If $\hat{r} \notin R$, then it is included and the game is solved again, otherwise the exact solution of the minmax problem is found. In the above formulation, only the integrality constraints on $y_t$ can be relaxed still obtaining a feasible optimal solution.

\begin{theorem} Deciding if there is at least a joint covering route where a subset of targets $T'$ can be covered with at most $k$ single--resource routes is $\mathcal{NP}$--hard.
\end{theorem}
\vspace{-0.1cm}
{\em Proof.} A simple reduction from SET--COVER can be given by map the universe on $T'$, set $m=k$, and define $R_1 = R_2 \ldots = R_m$ a set of one route for each set $S_j$ in SET--COVER where the targets covered by the route correspond to the elements of $S_j$. It is easy to see that SET--COVER admits a cover with at most $k$ sets if we can cover $T'$ with no more that $k$ single--resource routes (one for each resource). \hfill $\Box$

In practice, we empirically assessed that the following heuristic method gives acceptable results. We solve the linear relaxation (on $x_{ir}$ variables) an then sample a route $r$ for each resource $r$ with probability $x_{ir}$. We iterate this double--LP scheme until convergence, namely when the best response computed by the above program is pure and already in \Def's support at the equilibrium. 
 
\subsection{Partial coordination SRO (PC--SRO)}
Partial coordination models those situations where \Def~can act as central planner but cannot communicate with each resource to prescribe a joint action. Another way to see this is to consider each resource as a separate player that has interest in cooperating with others but that cannot correlate with them. Real scenarios falling in this scope can be characterized by resources for which a communication with the control unit is not available (e.g., patrolling units operating in stealth mode).

Team games~\cite{von1997team} are a suitable model to describe our security game under this coordination scheme. We define a $m+1$ game where resource players $\mathcal{D}_1, \ldots, \mathcal{D}_m$ compete together against \Att. Each resource strategy is defined as $\sigma^{\mathcal{D}}_i: R_i \rightarrow [0,1]$, where $\sigma^{\mathcal{D}}_i(r_i)$ denotes the probability of having resource $i$ following covering route $r_i \in R_i$. A game outcome is again defined with $(r,t)$ ($r$ is a joint covering route) where each $\mathcal{D}_i$ receives the same utility $U_{\mathcal{D}}$ as previously defined. The proper solution concept of this game is the team maxmin strategy whose computation can be described with the following non--linear program:

\begin{scriptsize}
\begin{align*}
\min\limits_{\sigma^{\mathcal{D}}_1 \ldots \sigma^{\mathcal{D}}_m} v & 	\quad \text{s.t.} \\
v - \pi(t) \prod_{i=1}^{m}\big( 1 - \sum_{r \in R_i} I(r,t)\sigma^{\mathcal{D}}_i(r)\big) \geq 0 & \quad \forall t \in T \\
\sum_{r \in R_i}\sigma^{\mathcal{D}}_i(r)= 1 & \quad \forall i \in \{1 \ldots m\} \\
\sigma^{\mathcal{D}}_i(r) \geq 0 & \quad \forall i \in \{1 \ldots m\}, r \in R_i 
\end{align*}
\end{scriptsize}

Notice that, the minmax value with three or more players can be irrational and not exactly computable even when the payoffs can assume only three different integer values. Approximating it with additive gap is, in general, hard. More precisely, approximating the minmax value of 2 players against 1 player within an additive error of $\frac{1}{3z^2}$ where $z$ is the number of actions of each player is $\mathcal{NP}$--hard even with binary payoffs~\cite{DBLP:journals/geb/BorgsCIKMP10}. \cite{DBLP:conf/wine/HansenHMS08} provides a quasi--polynomial algorithm with complexity $O(z^{l \frac{\log(z)}{\epsilon^2}})$ approximating with additive error $\epsilon$ the minmax value with $l$ players. It requires, for each player, to enumerate all the possible supports composed of $\log(n)$ actions. However, such methods become impractical for the problem we need to solve even for toy instances and non--negligible $\epsilon$ (e.g., with 10 actions, 2 resources, and $\epsilon = 0.5$, the number of needed iterations is of the order of $10^{18}$).

Thus, to cope with the resolution of this problem we propose the following iterative method. Consider a starting strategy profile $\hat{\sigma}^{\mathcal{D}}_1, \ldots, \hat{\sigma}^{\mathcal{D}}_m$ with the corresponding game value $\hat{\nu}$. Define $\nu_i$ as the value returned by the linear program obtained from the above formulation by setting $\sigma^{\mathcal{D}}_j = \hat{\sigma}^{\mathcal{D}}_j$ for any $j \neq i$. We then compute $i' = \arg\max_{i} \nu_i$ solving $m$ linear programs, set $\hat{\sigma}^{\mathcal{D}}_{i'} = \sigma^{\mathcal{D}}_{i'}$ and repeat until convergence. Notice that, at each iteration, the value of the game increases (non--strictly) monotonically. Finally, once convergence achieved, we have a random restart generating a new strategy profile for $\mathcal{D}$. Let us notice that, in the case the number of routes per resource is large, a row--generation approach similar to that described in the previous section can be adopted.

\subsection{No coordination SRO (NC--SRO)}\label{subsec:nc}
When no coordination is allowed between resources we are ruling out not only strategy correlation but also joint planning. In this scenario resources are unaware of each other and act as single players against \Att. This case is modeled with $m$ independent single resource signal response games (as defined in~\cite{basilico2015security}) where the game associated with resource $i$ has $p_i$ as starting vertex and is played on the restricted set of targets $T_i = \{t \in T \mid \omega_{p_i,t} \le d(t)\}$. For each game $i$, the maxmin strategy is computed taking $R_i$ as the action space of $\mathcal{D}_i$. However, although the defensive resources are completely non--coordinated, \Att~can observe the strategy of each resource and play at best. Thus, given the $m$ resource allocation strategies obtained in this way, we can compute the game value as $1 - \max_{t \in T}\prod_{i=1}^{m}\big(1 - \sum_{r_i \in R_i}\sigma^{\mathcal{D}}_i(r_i)I(r,t)\big)\pi(t)$.

%% file: 05-covering_placement.tex
\section{Overall resolution approach}
\begin{figure}[htbp]
\centering{\includegraphics[width=0.7\columnwidth]{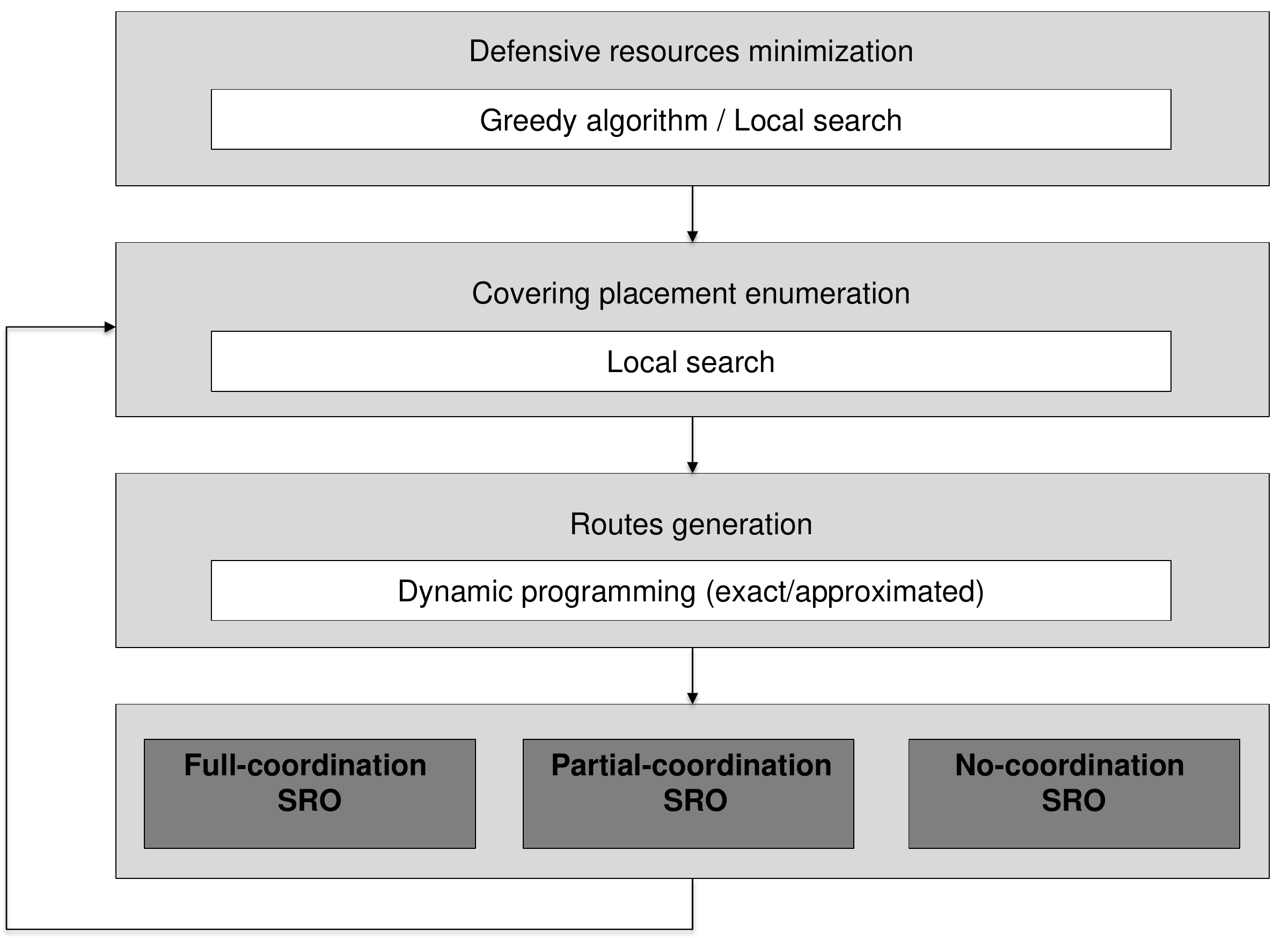}}
\caption{Overview of the proposed resolution approach.}\label{fig:overview}
\end{figure}

The contributions we presented are organized in the resolution approach sketched in Figure~\ref{fig:overview}. We start by tackling the problem of finding the minimum--size resource placement by solving the associated SET--COVER formulation (or our polynomial algorithm with trees and cycles). If the problem cannot be solved exactly in a short time, we adopt the greedy approximation algorithm of~\cite{chvatal1979greedy} and then apply a simple local search to improve the greedy solution. Once this step is terminated, we have fixed a number of resources $m$ and we address the problem of finding the best allocation strategy for them. As previously stated, in absence of false negatives and false positives the best allocation strategy when no signal is raised is to statically place the $m$ resources in the best covering placement. Such placement is defined as the one from which the signal response game yields the maximum expected utility. To deal with this, we resort again to a simple variation of the local search procedure~\cite{musliu2006local} to enumerate covering placements of exactly $m$ resources. For each considered covering placement, we compute sets $R_i$ for each resource $i$ as mentioned in the previous section and then we run the signal response oracles we introduced. 

%% file: 06-experiments.tex
\section{Experimental evaluations}
\label{sec:experimental_evaluation}
To evaluate our algorithms we implemented a random instance generator that leverages some domain knowledge we gathered by discussing with representatives of the Italian State Police. From such process, we were able to develop a tool for randomly generating patrolling instances that could represent realistic urban environments, such as streets, squares or districts, capturing scenarios with police patrolling units placed in police stations. In the graphs, all the vertices are targets, $|T| \in \{20, 40, 60, 80, 100, 120\}$, edges are unitary, the average indegree of each vertex is 3, and penetration times have been set according to the size of the instances as follows: for $|T| \in \{20, 40\}, d(t) = 3$, for $|T| \in \{60, 80\}, d(t) = 4$ and finally for $|T| \in \{100, 120\}, d(t) = 5$. There is one single signal covering all the targets, corresponding to the worst case in computational terms. Since the algorithms are anytime, the whole resolution process depicted in Figure~\ref{fig:overview} is given a timeout of $60$ minutes, thus implicitly posing a limit on the maximum size of instances that can be solved. All the numerical results we report are averaged on $50$ instances for each $|T|$. For the instances employed here, we used the exact method of~\cite{bdg2015advpatr} to generate the covering routes, requiring a compute time comparable to the approximation algorithm. We run experiments in MATLAB on a UNIX computer with 2.3GHz CPU and 128GB RAM. 

{\bf Resources and dispositions} We initially evaluate the performance of the algorithms to find the best covering placement. As we discussed in Section~\ref{sec:special_instances}, we tackle this problem by solving an associated SET--COVER problem for which we have an exact method based on an ILP and an approximation one combining a greedy heuristic with a local search procedure. The ILP--based method scales up to instances with $500$ vertices when evaluated in standalone runs and up to $120$ vertices in the scope of the whole resolution process (in this last case a portion of time is necessarily spent for the other subtasks listed in Figure~\ref{fig:overview}). The approximation algorithm achieves an average error $<5\%$ up to 500 targets, suggesting that it can provide good suboptimal results even for settings with more than 500 targets. In the following analysis, we use the ILP--based method. 

\begin{figure}[!htbp]
\begin{scriptsize}
\centering
\subfigure[Resources and dispositions]
  {\includegraphics[width=0.23\textwidth]{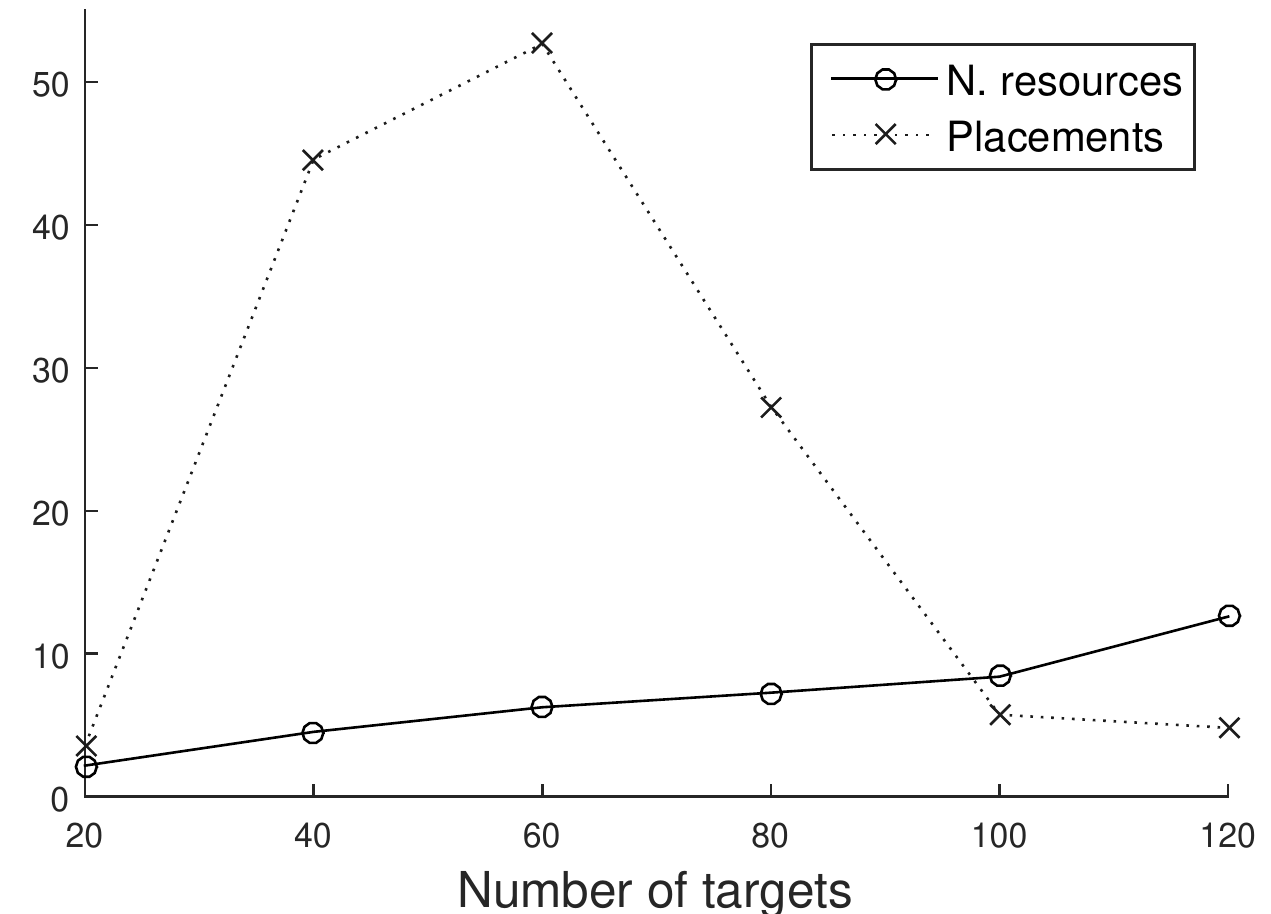}\label{fig:assignments}}
\subfigure[Overlap degree]
 {\includegraphics[width=0.23\textwidth]{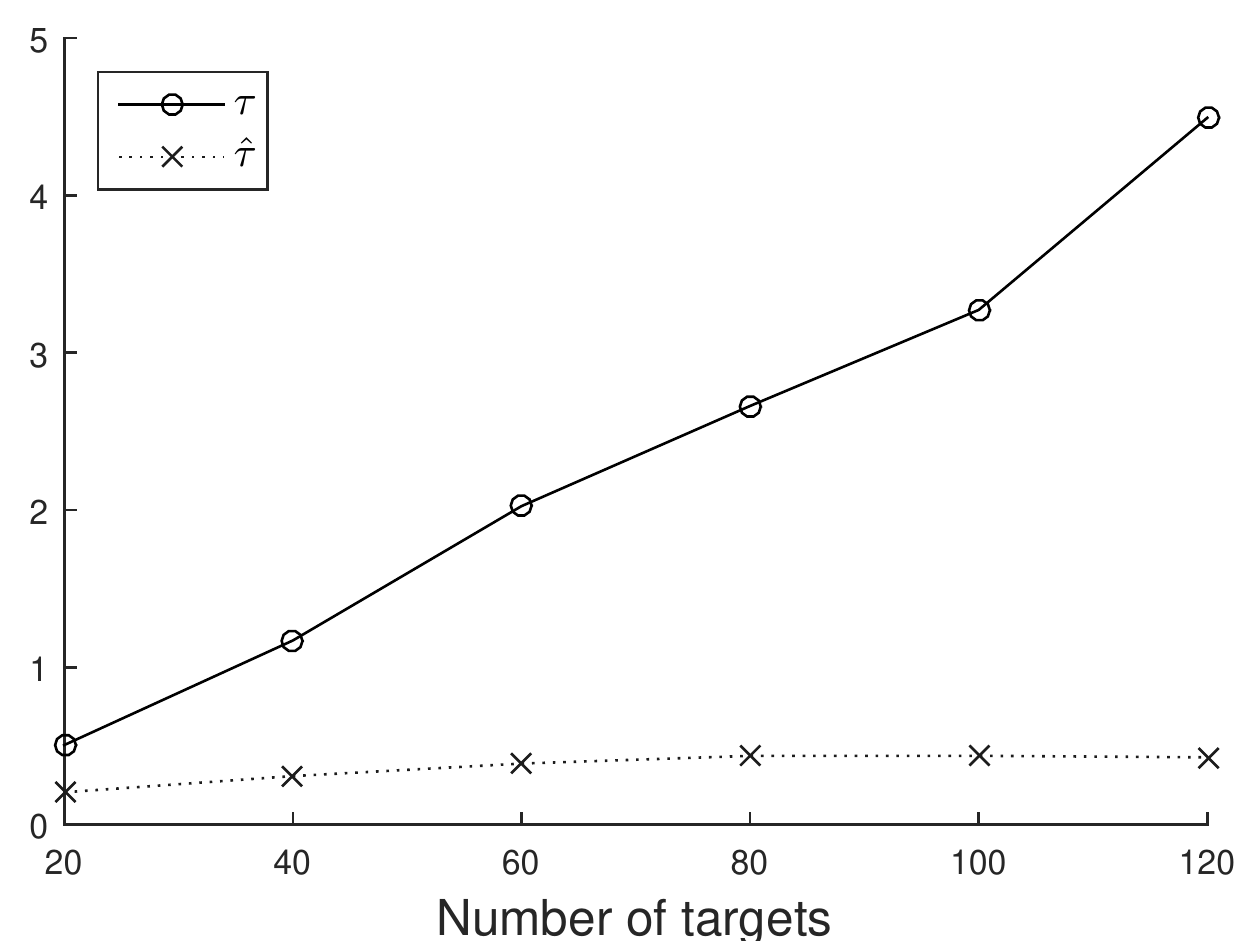}\label{fig:overlap}}
\end{scriptsize}
\caption{Analysis of the instances before an attack.}\label{fig:instances}
\end{figure}

First, we focus on the characteristics exhibited by our experimental settings in terms of number of resources needed for the covering placement and the degree of overlap over the targets covered by the resources, see Figure~\ref{fig:instances}. 

We observe that the average number of resources is linear in the number of targets (see Figure~\ref{fig:assignments}), which is reasonable in real--world scenarios.

To quantify the overlap degree induced by a given covering placement $P=\{p_1, p_2, \ldots, p_m\}$ we define two complementary indicators. We recall that $T_i$ is the subset of targets that can be covered (reached by their deadlines) from vertex $p_i$ (see Section~\ref{subsec:nc}). Then the following quantity counts the number of {\em extra coverings} induced by $P$: $\eta = \sum_{i=1}^{m}|T(p_i)| - |T|$. Notice that $0 \le \eta \le (|T|-m)(m-1)$ getting its maximum value when each resource covers a different subset of $|T|-m+1$ targets. We denote with $\tau = \frac{\eta}{|T|}$ the average overlap per target and $\hat{\tau} = \frac{\eta}{(|T|-m)(m-1)}$ the normalized overlap. Figure~\ref{fig:overlap} shows how these two indicators vary with the number of targets. The index $\tau$ grows linearly in the number of targets due to the linear growth of the number of resources while $\hat{\tau}$ has a slower growth since for any resource $i$ the number of covered targets $|T(p_i)|$ is usually not as big as $|T|$.

Now we focus on the covering placement enumeration phase and we evaluate the number of covering placements that our algorithm is able to consider within a $60$ minutes deadline. Referring again to Figure~\ref{fig:overview}, notice that each additionally considered placement requires to compute the covering routes sets and to run one of the three SROs. This time is dominated by the covering route computation phase, being the time required by each oracle never exceeding $1$ minute. The number of generated covering placements is reported in Figure~\ref{fig:assignments}. The curve grows reaching its maximum for $|T| = 60$ and then decreases. Indeed, when the size of the problem is small, the algorithm terminates before the deadline, returning few placements. On the other side, when the size of the problem is large, due to the time limit and the huge amount of time required by the computation of the covering routes, the time to compute new placements lowers significantly and thus the curve decreases. We remark that we are able to solve large instances up to $120$ targets.

\begin{figure}[!htbp]
\begin{scriptsize}
\centering
 \subfigure[Utility ratios]
 {\includegraphics[width=0.23\textwidth]{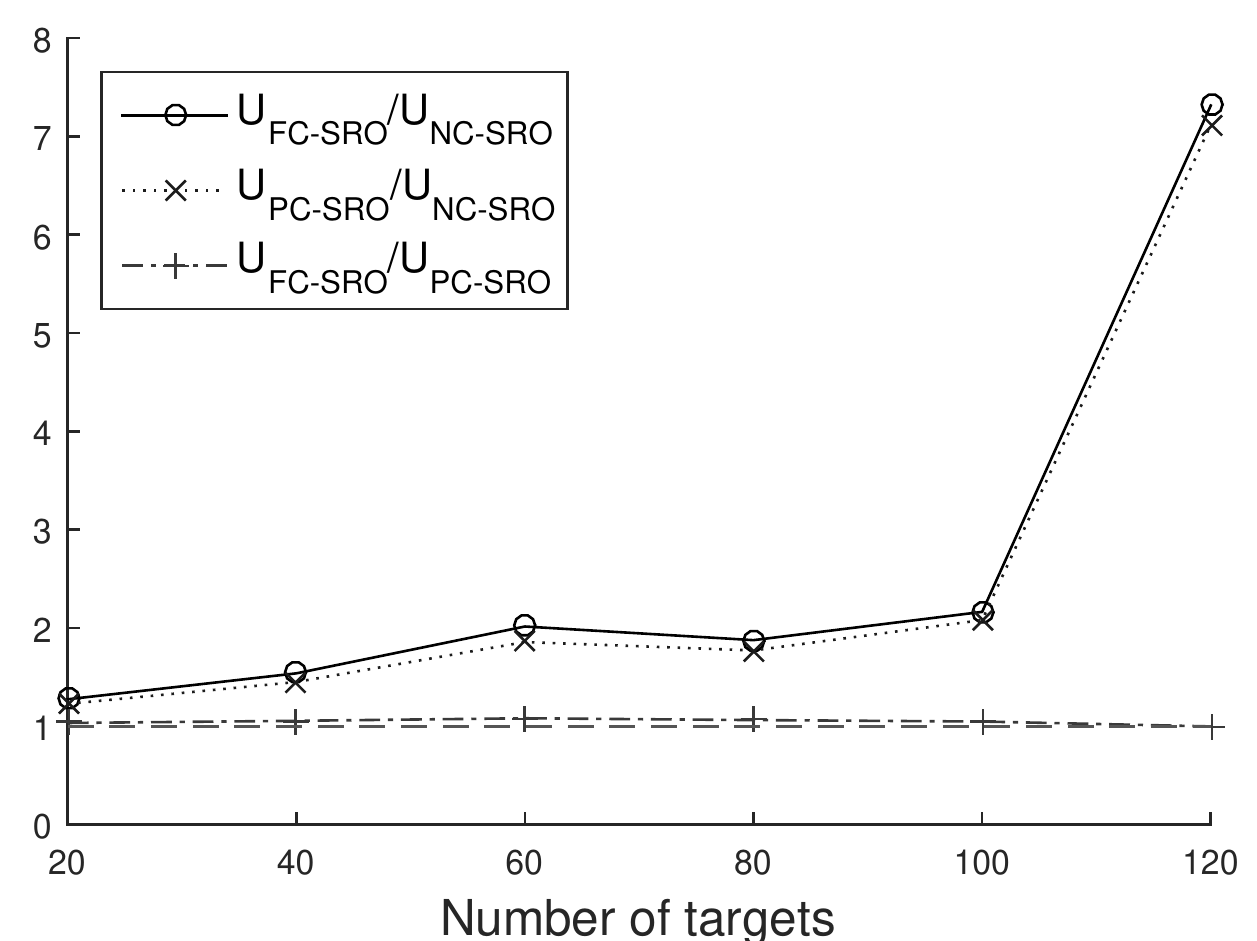}\label{fig:utility}}
  \subfigure[Time ratios]
  {\includegraphics[width=0.23\textwidth]{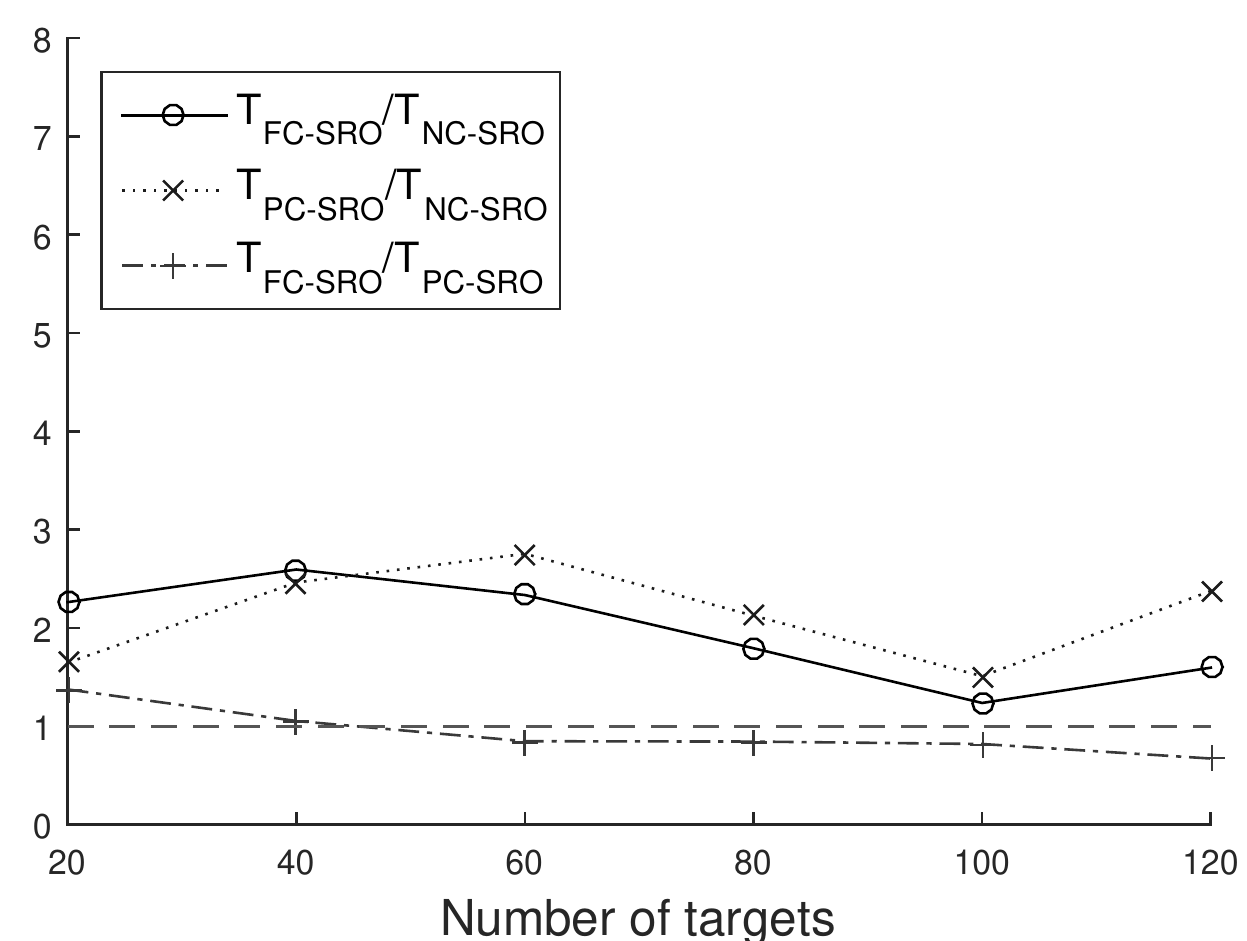}\label{fig:time}}
\end{scriptsize}
\caption{Utility and time ratios of the three SROs.}\label{fig:oracles}
\end{figure}

{\bf SROs quality and performance}
Here we compare the performance of the three SROs, both in terms of utility and compute time. We exploit the solution returned by the NC--SRO to initialize the PC--SRO and the FC--SRO (in this last case as input for the row--generation algorithm). We evaluate the PC--SRO both with $0$ and $10$ random restarts observing that the improvement is not significant ($<1\%$) and confirming that choosing the solution of the NC--SRO as the starting point is appropriate. For the FC--SRO, we evaluate both the exact and the approximation row--generation algorithm, finding that the average difference between the two objective functions is negligible up to $120$ targets ($<1\%$), suggesting that the linear relaxation works very well in practice. In the results we report below we use $0$ random restarts for PC--SRO and the exact row--generation algorithm for FC--SRO (returning thus the optimal strategy). Figure~\ref{fig:utility} shows that enabling coordination among the resources gives a significant burst in the growth of the utility w.r.t. the number of targets, especially after 100 targets. As expected, FC--SRO guarantees the highest utility. Interestingly, the performance of PC--SRO is very close to that one of FC--SRO, suggesting that in our settings the price of partial coordination is low. Figure~\ref{fig:time} shows the time ratio among the three SROs. The absolute times are all very low (usually on average $<20$ seconds even for $120$ targets). We observe that both FC--SRO and PC--SRO present a linear growth in time w.r.t. NC--SRO but FC--SRO requires less time. This suggests that the row--generation of FC--SRO works very well, generating a small set of joint routes. Hence, FC--SRO is the best oracle in terms of tradeoff between utility and compute time, when full coordination is possible.

\begin{figure}[!htbp]
\begin{scriptsize}
\centering
 \subfigure[Utility ratios]
 {\includegraphics[width=0.23\textwidth]{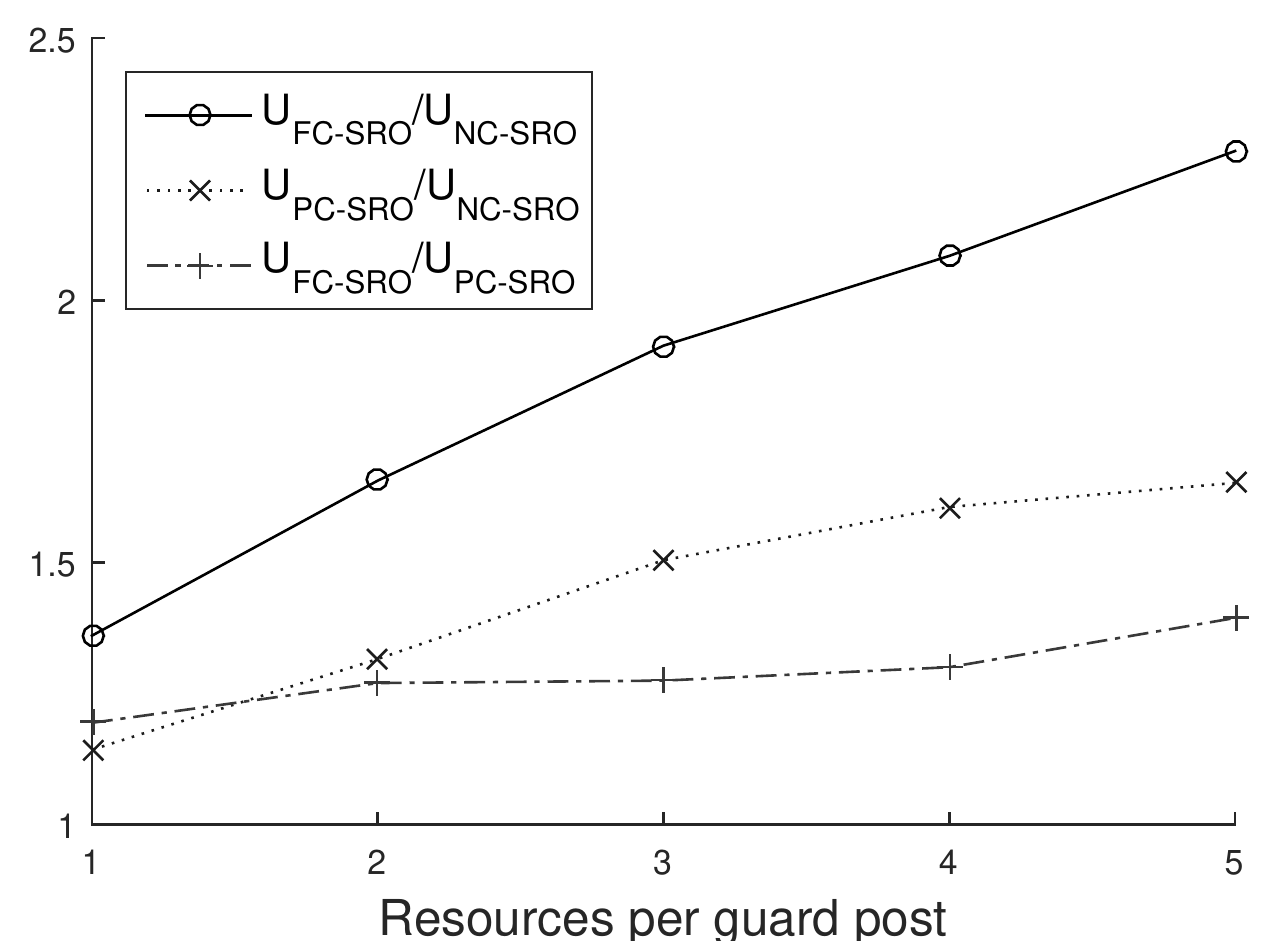}\label{fig:more_utility}}
  \subfigure[Time ratios]
  {\includegraphics[width=0.23\textwidth]{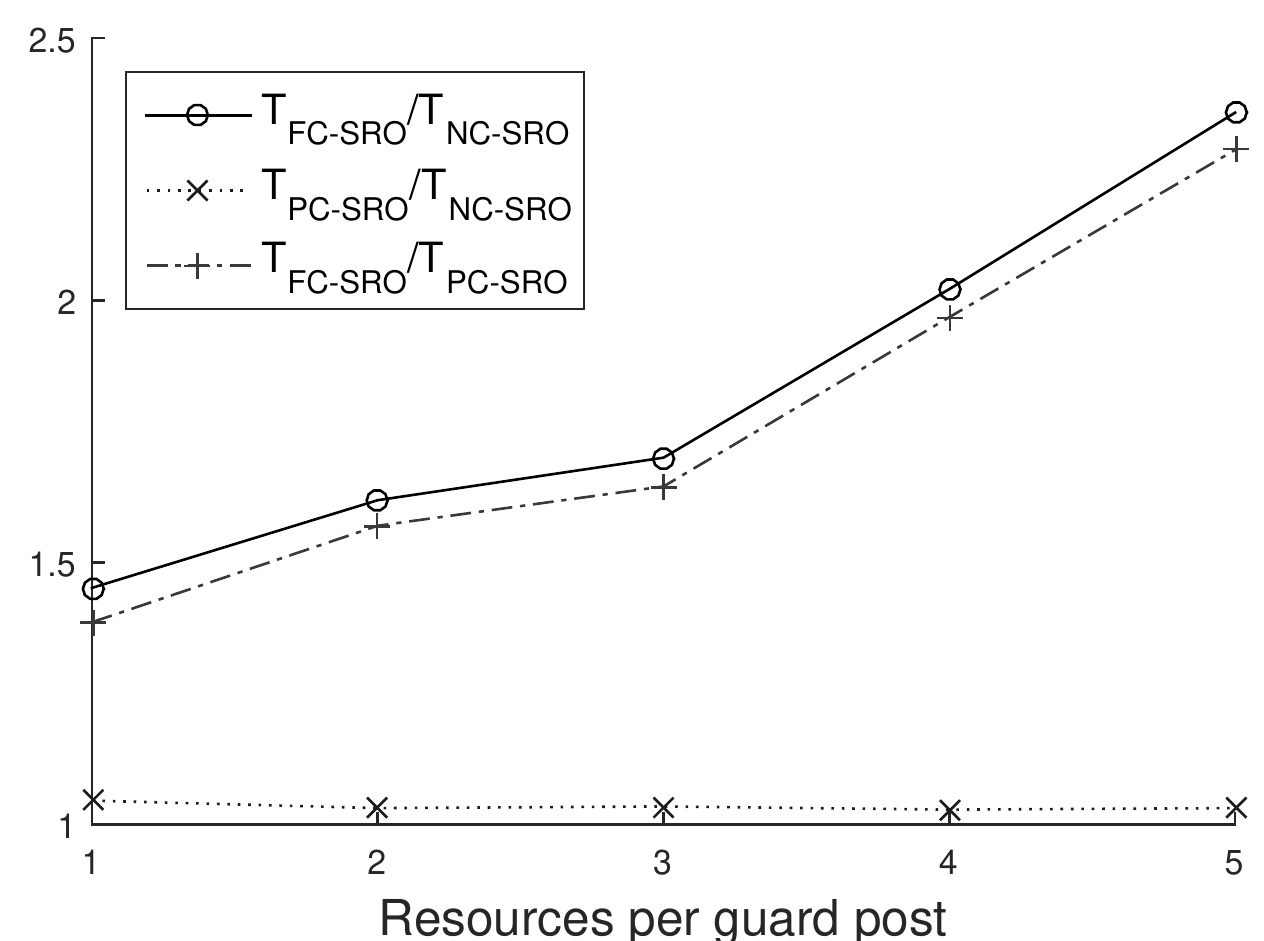}\label{fig:more_time}}
\end{scriptsize}
\caption{Utility and time ratios with more resources per guard post.}\label{fig:more}
\end{figure}

Finally, we evaluate the performance of the SROs as the number of resources varies. We conduct experiments on graphs with $|T| = 80$ (instances with more targets require excessive computational costs) in which the best $|P|$ = 4. Then, we consider the resources as guard posts and we assign up to $5$ mobile patrollers to each guard post. The FC--SRO needs more time w.r.t. PC--SRO, as it can be seen in Figure~\ref{fig:more_time}, but this difference is very small. Observing Figure~\ref{fig:more_utility}, we note that also in this case enabling coordination gives more utility. Thus, we conclude that the FC--SRO is the best performing method, providing the highest utility and requiring only slightly more time than the PC--SRO. On the other side, the PC--SRO is a viable good solution since, even without coordination, it still provides good utility results.

\begin{figure}[!htbp]
\begin{scriptsize}
\centering
 \subfigure
 {\includegraphics[width=0.23\textwidth]{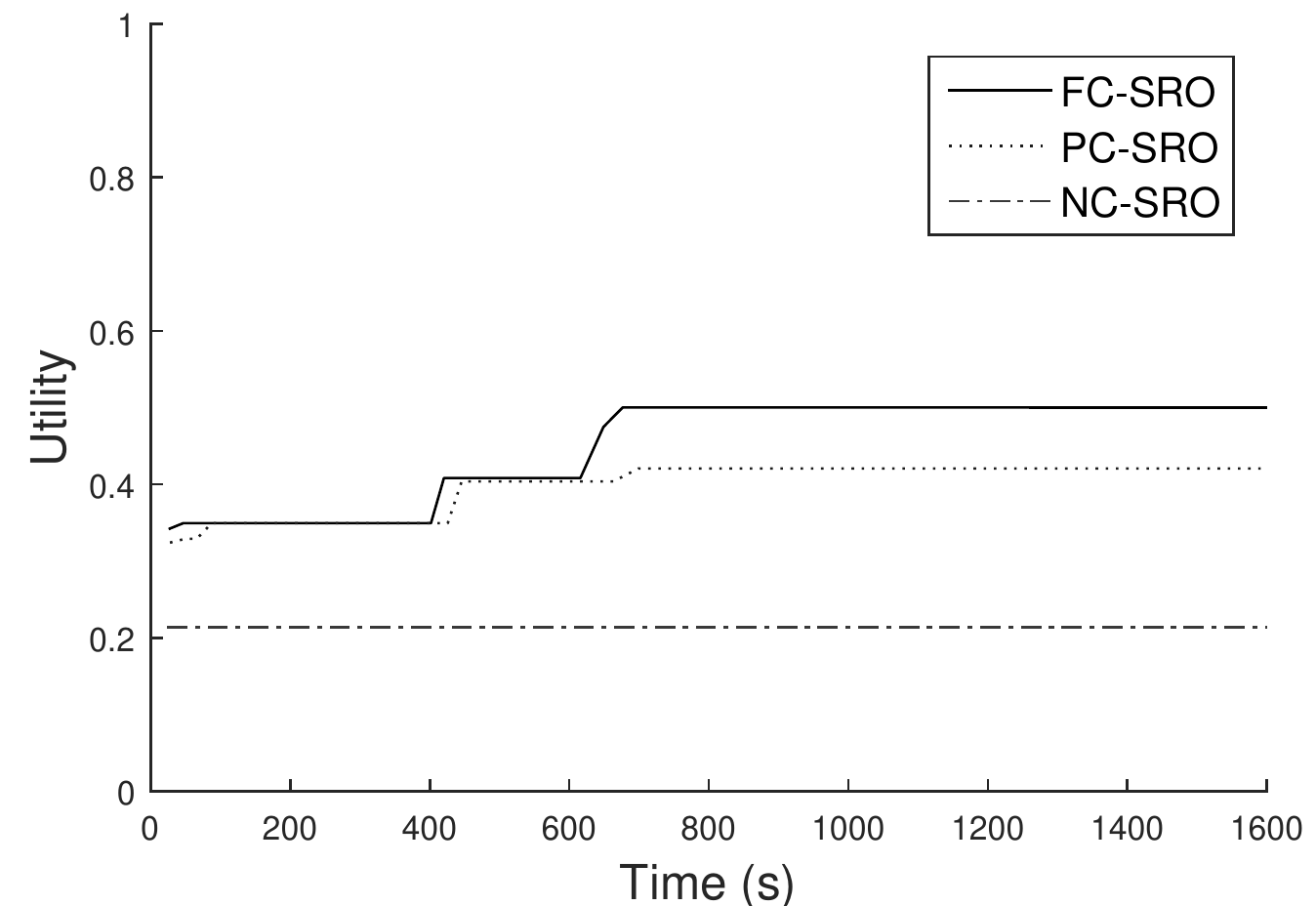}\label{fig:utility_in_time_2}}
  \subfigure
  {\includegraphics[width=0.23\textwidth]{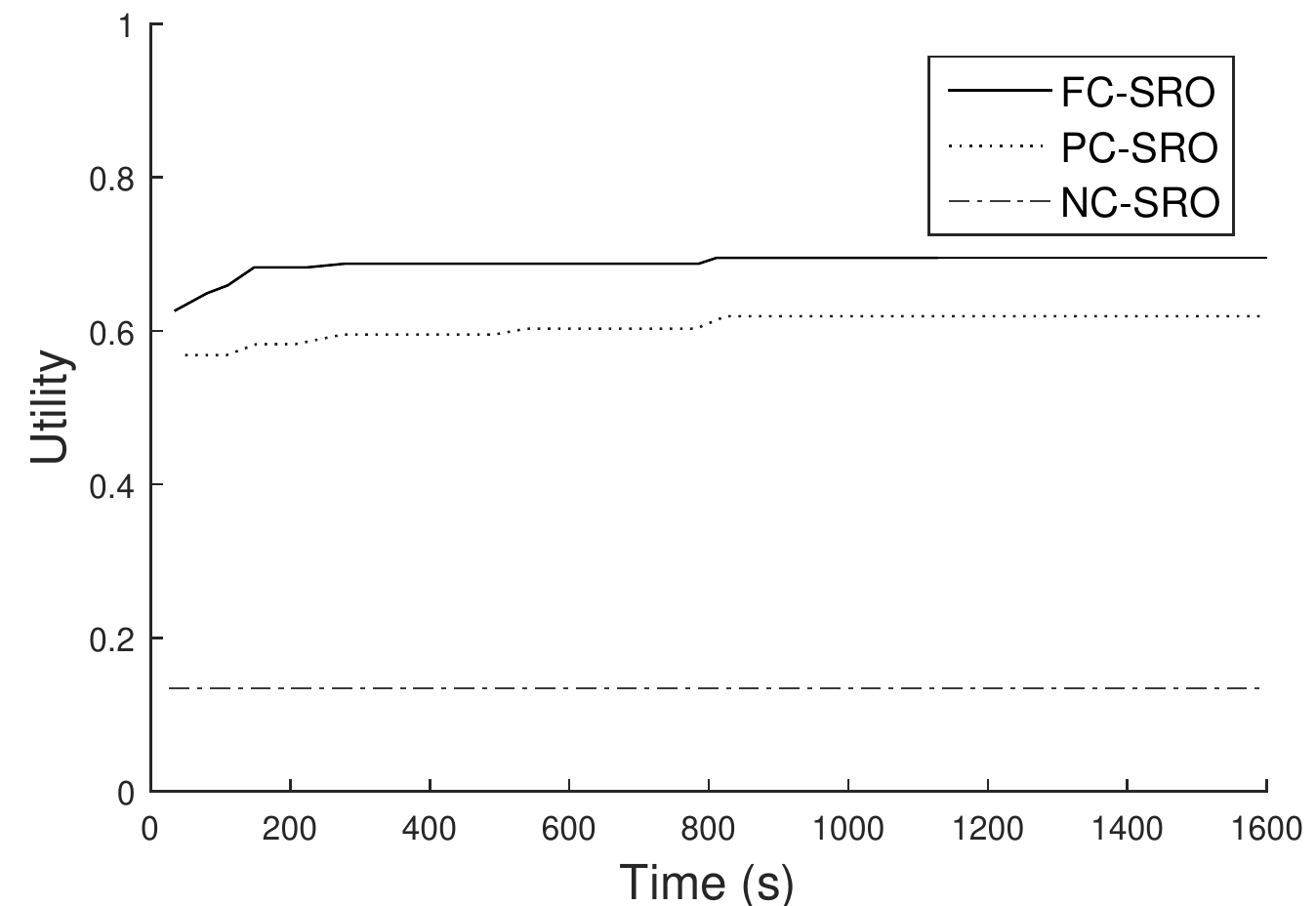}\label{fig:utility_in_time_1}}
\end{scriptsize}
\caption{Utility trend in time.}\label{fig:utility_in_time}
\end{figure}

{\bf Utility trend in time}
Now we focus on the evolution of the utility in time, after different placements have been enumerated and evaluated. Figure~\ref{fig:utility_in_time} shows two instances with $|T| = 60$ where we compare the performances of the three oracles with one resource per guard post. While NC--SRO is quite constant even though several placements are evaluated, FC--SRO and PC--SRO utilities increase, with the former always preceding the latter due to its lower computational time. 

%% file: 07-conclusions.tex
\section{Conclusions and future research}\label{sec:conclusions}
In this work, we considered a security game with the presence of a spatially imperfect alarm system and we addressed the novel generalization towards settings in which the Defender can control multiple mobile resources. We proposed a resolution approach for dealing with the new algorithmic problems that such generalization introduces. First, we addressed the problem of computing the minimum number of resources required in a given setting and then we tackled the determination of the best signal response strategy under different resource coordination schemes. 

Future direction of research will involve adaptations of our algorithms to cases in which the number of resources available to $\mathcal{D}$ is greater than the ones required for the minimum covering placement. 
In addition, we plan to work on the model by allowing the presence of false positives and missed detection in the alarm systems as well as the presence of multiple resource from the attacker side. These extensions are naturally driven by real--world requirements to which security games must comply.